\documentclass[conference]{IEEEtran}
\IEEEoverridecommandlockouts

\usepackage{cite}
\usepackage{amsmath,amssymb,amsfonts}
\usepackage{graphicx}
\usepackage{textcomp}
\usepackage{xcolor}
\usepackage{booktabs}
\usepackage{multirow}
\usepackage{array}
\usepackage[colorlinks=true, urlcolor=blue, linkcolor=black, citecolor=black]{hyperref}
\usepackage{float}
\usepackage{subcaption}
\usepackage{enumitem}
\usepackage{algorithm}
\usepackage{microtype}
\usepackage{balance}
\usepackage{algpseudocode}

\def\BibTeX{{\rm B\kern-.05em{\sc i\kern-.025em b}\kern-.08em
T\kern-.1667em\lower.7ex\hbox{E}\kern-.125emX}}

\begin{document}

\title{Reactive 3D Motion Planning for a Franka Arm via Star-World Workspace Reshaping}

\author{
\IEEEauthorblockN{Gia Dcosta, Saayuj Deshpande, and Samhitha Vedire}
\IEEEauthorblockA{University of Pennsylvania, Philadelphia, PA, USA}
}


\maketitle

\begin{abstract}
Safety inflation can cause nearby obstacles to overlap, violating the disjoint-obstacle assumptions used by many modulation-based reactive planners. We investigate Star-World workspace reshaping for three-dimensional reactive control of a Franka Emika Panda manipulator. At each update, intersecting inflated obstacles are clustered and replaced by star-shaped proxies before a dynamical-system-based end-effector controller is evaluated. A null-space artificial-potential-field term provides complementary arm-body avoidance. We compare reshaped and unreshaped obstacle representations in six PyBullet scenarios using goal attainment, path-length ratio, and computation time. In this preliminary 12-trial evaluation, reshaping reaches the goal in five of six scenarios, compared with four of six for the unreshaped baseline. It resolves the canonical overlapping-wall case and requires 0.68--8.70\,ms per workspace update for scenes containing one to seven obstacles. However, it also increases path length, produces near-equilibria in two cases, and closes a navigable corridor through over-aggressive merging. These results show both the promise and the practical limitations of transferring Star-World guarantees from workspace geometry to a redundant manipulator controlled through inverse kinematics.
\end{abstract}


\section{Introduction}

Robotic manipulators operating in real-world environments must continuously react
to obstacles while still reaching desired goals safely and efficiently. This
problem becomes especially important in cluttered workspaces where nearby
objects, narrow passages, and uncertain geometry can significantly affect robot
motion.

Traditional global trajectory planning approaches compute a path ahead of time
using a fixed environment model. While these methods can generate optimal paths,
they are often computationally expensive and less adaptable to changing
environments. Reactive planning methods instead continuously update robot motion
online based on the current robot state and nearby obstacle information. This
allows robots to quickly respond to changes in the workspace while maintaining
smooth motion toward a goal.

Dynamical-system-based (DS) motion planning is particularly attractive for
reactive manipulation because it generates smooth and continuous vector fields.
A nominal DS drives the robot toward the goal while modulation techniques modify
the velocity field near obstacles to avoid collisions. Unlike trajectory
optimization methods that repeatedly solve optimization problems online,
DS-based methods are lightweight and naturally suited for continuous reactive
control.

However, many modulation-based planners assume that the workspace forms a
disjoint star world in which obstacles are star-shaped and non-overlapping. In
realistic environments this assumption is frequently violated, and the planner
may generate highly conservative motion, oscillatory behavior, or fail to
converge entirely.

This problem becomes even more challenging for robotic manipulators such as the
Franka Panda arm. The robot must navigate through cluttered three-dimensional
environments while accounting for both end-effector motion and arm-body
collision avoidance.

The goal of this project is to investigate whether Star-World workspace reshaping
can improve reactive planning performance in cluttered 3D scenes. Instead of
directly planning around overlapping inflated obstacles, intersecting obstacles
are grouped into clusters and reshaped into cleaner star-shaped obstacle regions
before planning.

Our implementation extends the Star-World reshaping ideas proposed by Dahlin and
Karayiannidis~\cite{dahlin2023} into a 3D manipulator setting. We evaluate the
proposed approach on a Franka Panda arm using success rate, path length
ratio, and computational runtime.

\subsection{Motivation and Contributions}

\begin{figure}[h]
    \centering
    \includegraphics[width=0.9\linewidth]{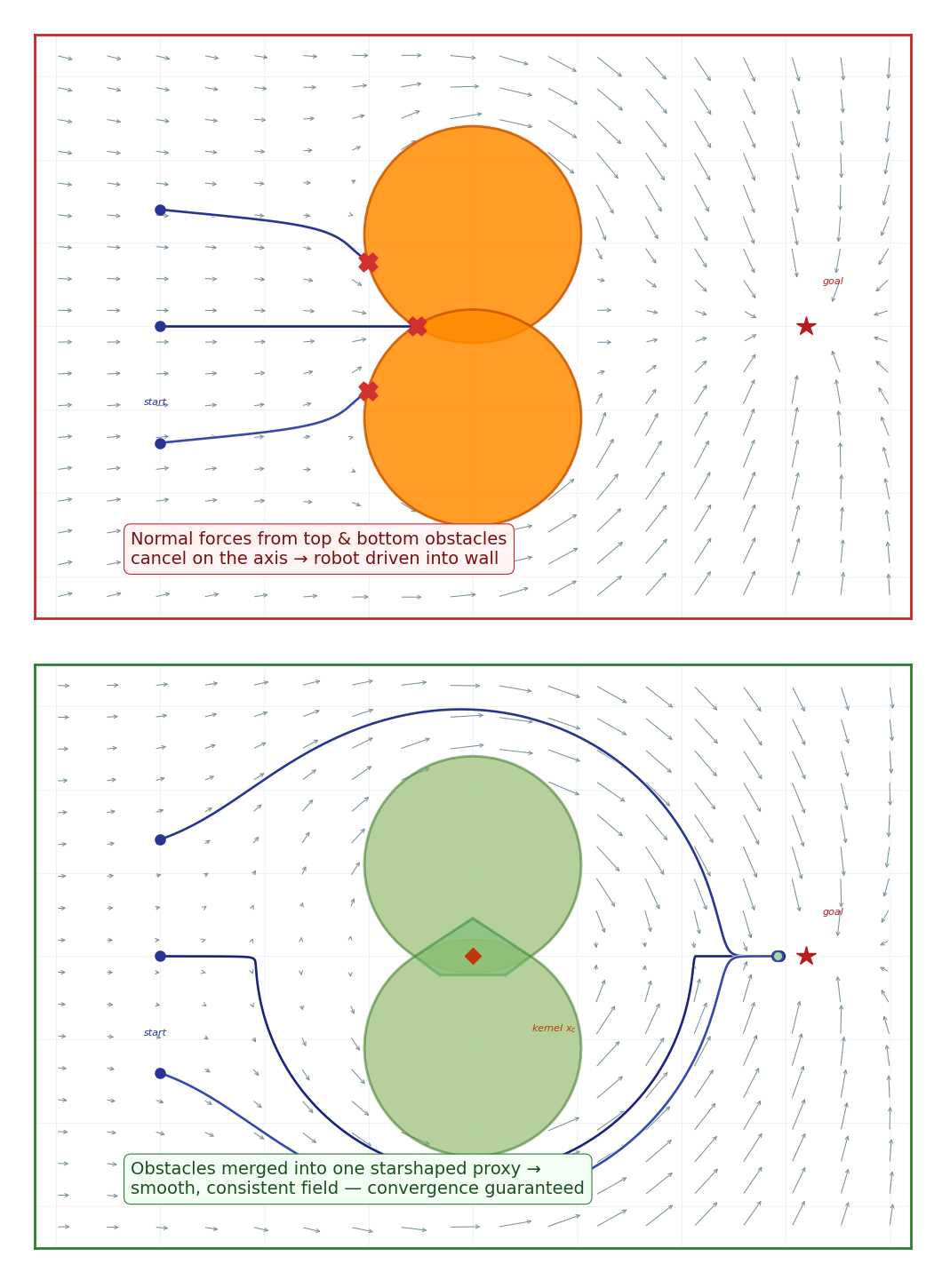}
    \caption{\small 2-D illustration of the core failure mode. \textit{Upper}: two
             overlapping circles modulated independently - repulsion vectors
             cancel on the approach axis, driving the robot into the wall.
             \textit{Lower}: Algorithm~2 merges both circles into one proxy;
             the offset kernel gives a coherent tangential deflection.}
    \label{fig:2d_comparison}
\end{figure}

Reactive obstacle avoidance becomes increasingly difficult when multiple inflated
obstacles overlap or come very close together. In standard modulation-based
planners~\cite{huber2019avoidance, huber2022avoiding}, each obstacle
independently contributes a modulation matrix to the velocity field. When
several inflated obstacles intersect, the planner receives conflicting local
obstacle information that can distort or cancel the reactive velocity
field. This is illustrated in Fig.~\ref{fig:2d_comparison}: two overlapping circles
form an impassable wall, yet their independent repulsion vectors cancel exactly
on the approach axis, driving the robot directly into the barrier.  Star-World
reshaping merges both circles into a single proxy with an off-axis kernel,
producing a coherent tangential deflection that routes the robot around the
combined obstacle.

These effects become particularly severe in cluttered manipulation environments
where safety inflation is necessary to account for sensing uncertainty, robot
thickness, controller inaccuracies, and safety margins. Although inflation
improves safety, it also increases the likelihood of obstacle overlap and
violates the disjointness assumptions required for convergence guarantees of
modulation-based planners. In dense scenes, the baseline planner may:

\begin{itemize}
    \item Produce highly conservative or unsafe trajectories,
    \item Generate oscillatory behavior near obstacle intersections,
    \item Fail to converge to the goal due to conflicting modulation
          contributions,
    \item Exhibit poorly interpretable velocity field deformations.
\end{itemize}

Rather than independently modulating around several overlapping obstacles,
Algorithm~2 of Dahlin \& Karayiannidis~\cite{dahlin2023} clusters
intersecting obstacles, computes a star-shaped hull proxy per cluster with a
guaranteed kernel point, and presents the planner with a \emph{disjoint star
world}. This satisfies the structural preconditions for convergence. This work makes three contributions:
\begin{itemize}
    \item a three-dimensional implementation of Star-World proxy construction
          for reactive end-effector planning;
    \item an integrated controller that combines proxy-based DS modulation with
          null-space arm-body avoidance for a redundant manipulator; and
    \item a comparative simulation study that identifies both successful
          overlap handling and failure modes caused by conservative merging,
          joint limits, and heuristic kernel placement.
\end{itemize}

\section{Related Work}

\subsection{Dynamical-System-Based Motion Planning}

Dynamical systems are widely used in reactive robotics because they generate
smooth trajectories while remaining computationally lightweight. A nominal
attractor dynamical system is typically written as:

\begin{equation}
\dot{x}_{\text{nom}} = -K(x - x_g)
\end{equation}

where $x$ is the current robot position, $x_g$ is the goal position, and $K$
is a positive gain matrix. Obstacle avoidance is introduced through modulation
matrices that deform the nominal vector field near obstacles~\cite{khansari2012}.
Huber et al.~\cite{huber2019avoidance, huber2022avoiding} extended this
framework to guarantee convergence for convex and star-shaped obstacles,
provided the workspace satisfies a disjointness condition. These methods
preserve continuous online replanning while maintaining smooth trajectories,
and form the reactive controller used in this project.

\subsection{Artificial Potential Fields}

Artificial Potential Fields (APFs) are another popular reactive planning
approach~\cite{khatib1986}. Attractive forces pull the robot toward the goal
while repulsive forces push it away from obstacles. Although APFs are
computationally efficient, they may suffer from local minima and unstable
behavior in cluttered environments. In this work, APF-style repulsive terms
are used for arm-body collision avoidance while the end-effector follows the
DS-based modulation planner.

\subsection{Star-World Workspace Reshaping}

Dahlin and Karayiannidis~\cite{dahlin2023} proposed Star-World workspace
reshaping to address the disjointness requirement of modulation-based planners.
The paper introduces two key constructs: (1) the \emph{admissible kernel}, a
region from which an obstacle cluster can be seen as star-shaped while
excluding the robot and goal positions; and (2) the \emph{star-shaped hull with
specified kernel}, which merges intersecting obstacles into a single convex
proxy guaranteed to contain a valid kernel point. Together these allow
Algorithm~2 of~\cite{dahlin2023} to iteratively cluster overlapping obstacles
and produce a disjoint star world suitable for convergence-guaranteed reactive
planning. Our implementation extends this framework to the full 3-D end-effector space
of a Franka Panda arm.

\section{Methodology}

The complete planning pipeline consists of:
\begin{enumerate}[leftmargin=*]
    \item Scene modeling and obstacle inflation,
    \item Star-World reshaping (admissible kernel, kernel selection,
          star-shaped hull, re-clustering),
    \item Reactive DS planning with modulation,
    \item APF arm-body avoidance and null-space IK,
    \item Franka execution via IK position control.
\end{enumerate}

\begin{algorithm}[h]
\caption{Reactive Planning with Star-World Reshaping}
\small
\begin{algorithmic}[1]
\State Initialize Franka arm and obstacle set in PyBullet
\State Inflate obstacle radii by safety margin $\Delta r$
\While{goal not reached}
    \State \textbf{At 4 Hz:} query obstacle positions; form Ellipsoid objects
    \For{each obstacle cluster (Algorithm 2)}
        \State Compute admissible kernel (cone intersections / convex hull)
        \State Select kernel points $K$ (equilateral triangle / tetrahedron)
        \State Construct star-shaped hull proxy $\mathrm{SH}_K(O_{\mathcal{C}})$
        \State Re-cluster intersecting proxies; repeat if partition changed
    \EndFor
    \State \textbf{At 40 Hz:} compute nominal velocity $f_0 = (x_g - x)/\|x_g - x\|$
    \State Compute $\Gamma_i$ and modulation matrices $M_i$ for each proxy
    \State Apply proximity-weighted modulation: $\dot{x}_{\text{mod}} = M(x)f_0$
    \State Monitor all 9 arm links; compute APF repulsion when $\Gamma < 1.6$
    \State Project APF correction into null space: $\Delta q_{\text{ns}} = N\,\Delta q_{\text{corr}}$
    \State Solve IK seeded from current joints; apply null-space correction
    \State Command joints via PyBullet position control
\EndWhile
\end{algorithmic}
\end{algorithm}

\begin{figure*}[h]
    \centering
    \includegraphics[width=0.7\textwidth]{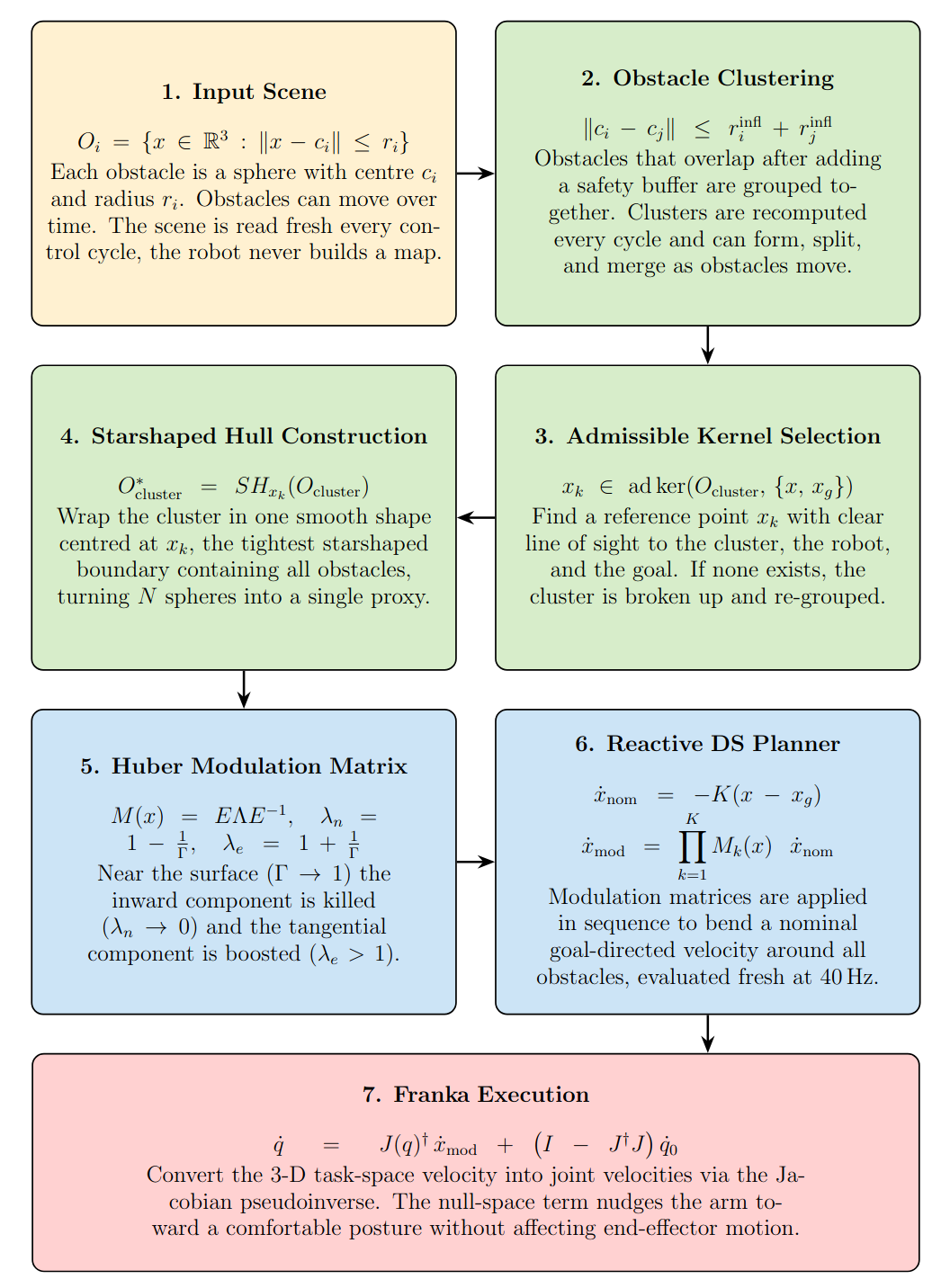}
    \caption{\small Complete reactive planning pipeline for Star-World workspace
             reshaping and reactive Franka arm control.}
    \label{fig:pipeline}
\end{figure*}

\subsection{Scene Modeling and Obstacle Inflation}

Obstacles are modeled as 3-D spheres represented as Ellipsoid objects
(degenerate case $Q = r^2 I$):

\begin{equation}
O_i = \{x \in \mathbb{R}^3 : \|x - c_i\| \le r_i\}
\end{equation}

where $c_i$ is the obstacle center and $r_i$ is the obstacle radius.  Each
obstacle is inflated by a safety margin $\Delta r$ before any planning step:

\begin{equation}
r_i^{\text{infl}} = r_i + \Delta r
\end{equation}

Inflation consolidates the robot's volume and sensing uncertainty into the
obstacle representation, allowing the planner to treat the end-effector as a
point.  A fixed margin of $\Delta r = 0.10$\,m is used in all experiments.
Obstacles are rendered as visual-only bodies in PyBullet (no physics collision
geometry) so that the simulator does not generate spurious contact forces that
would corrupt the EE trajectory.

\subsection{Star-World Reshaping}

Algorithm~2~\cite{dahlin2023} iteratively reshapes the inflated obstacles into
a disjoint star world.  The loop runs until the cluster partition stabilizes
(typically 1--2 iterations) or a safety cap is reached.

\subsubsection{Admissible Kernel}

For each obstacle cluster $\mathcal{C}$, the admissible kernel is the set of
candidate kernel points from which the union of obstacles appears star-shaped
while still excluding the robot and goal positions:

\begin{equation}
\mathrm{ad\,ker}(O_{\mathcal{C}},\,\{x,\,x_g\})
    = \bigcap_{i \in \mathcal{C}}\; \bigcap_{\bar{x} \in \{x,\,x_g\}}
      \mathrm{ad\,ker}(O_i,\,\{\bar{x}\})
\end{equation}

In 2-D each per-obstacle, per-excluded-point kernel is a cone fan polygon
(Eq.~8 of~\cite{dahlin2023}) whose apex lies at $\bar{x}$ and whose edges are
tangent lines to the obstacle.  Intersecting these cones via Sutherland--Hodgman
clipping gives the feasible kernel region for the cluster.  This cache is built
once outside the loop because the robot and goal positions are fixed within a
timestep.

If the admissible kernel is empty, the algorithm falls back to a convex
decomposition of each obstacle (Eq.~14 of~\cite{dahlin2023}).

\emph{3-D approximation.}
In 3-D the paper's cone-intersection procedure is approximated by the
convex hull of the cluster's surface samples.  This approximation is
sufficient in practice for sphere obstacles but does not carry the formal
guarantee of the 2-D construction: it is no longer strictly proven that
the robot position $x$ and goal $x_g$ are excluded from the resulting
proxy.  Tighter guarantees would require implementing the full 3-D cone
exclusion analogues of the 2-D admissible kernel.

\subsubsection{Kernel Selection}

Given the feasible kernel region, Algorithm~3 selects $n{+}1$ affinely
independent kernel points $K$ whose convex hull $\mathrm{CH}(K)$ lies strictly
inside the admissible kernel.  In 2-D, $K$ is an equilateral triangle centred
at the feasible region's centroid with side length proportional to the local
inscribed-circle radius; in 3-D a regular tetrahedron is used instead.
Temporal coherence is enforced by reusing the previous timestep's kernel
centroid when it remains inside the current feasible region, preventing the
robot from switching circumvention direction mid-trajectory.

\subsubsection{Star-shaped Hull and Re-clustering}

The star-shaped hull of cluster $\mathcal{C}$ with kernel $K$ is built as one
convex piece per obstacle (Property 4d of~\cite{dahlin2023}):

\begin{equation}
\mathrm{SH}_K\!\left(\bigcup_{i\in\mathcal{C}} O_i\right)
    = \bigcup_{i\in\mathcal{C}} \mathrm{SH}_K(O_i),
\qquad
\mathrm{SH}_K(O_i) = \mathrm{CH}(O_i \cup K)
\end{equation}

\begin{figure}[h]
    \centering
    \includegraphics[width=\linewidth]{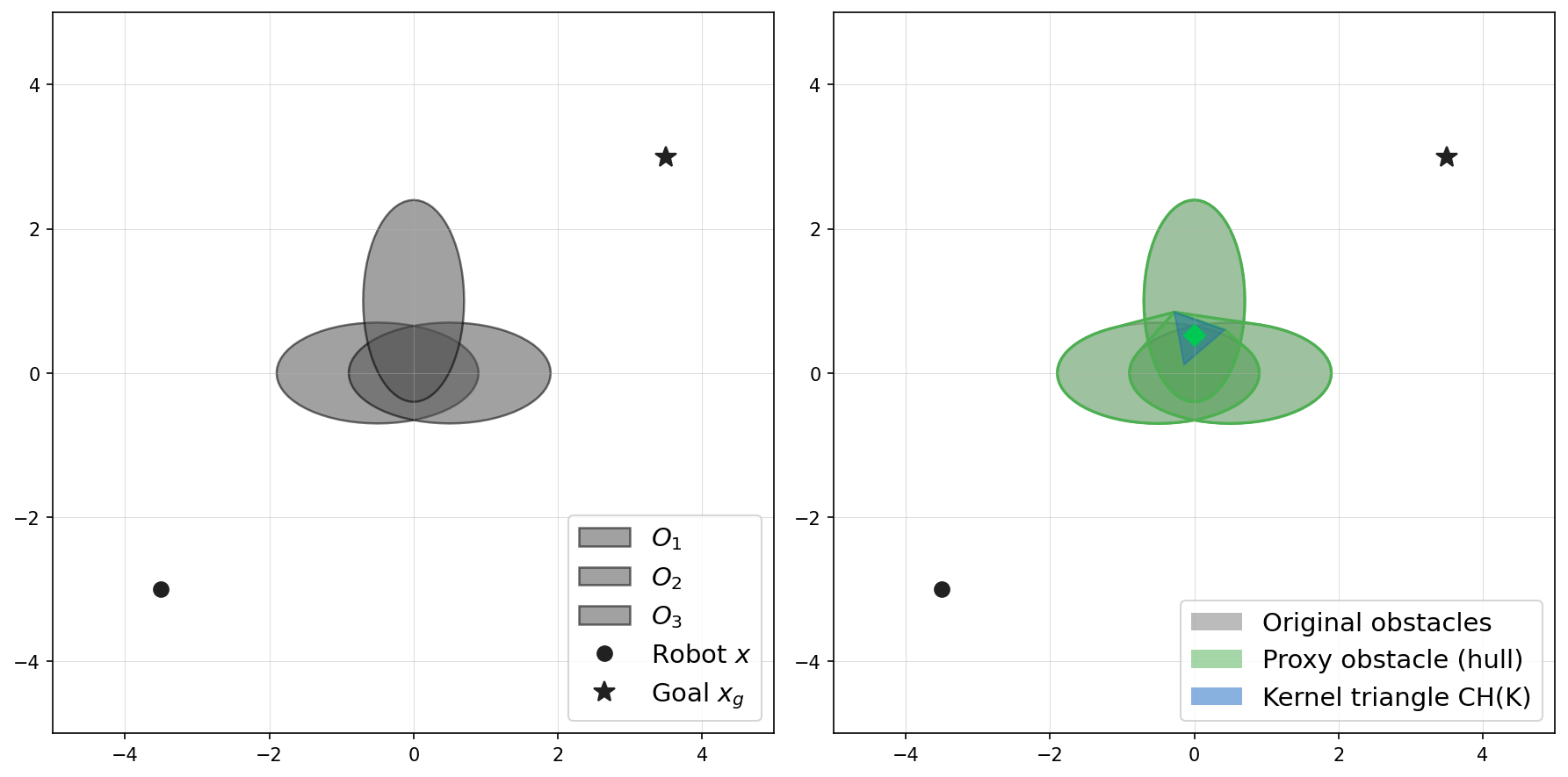}
    \caption{\small Star-World construction for three 2-D ellipse obstacles.
             Shaded regions show individual admissible kernel cones; the
             green triangle is the selected kernel $K$; coloured outlines
             are the per-obstacle star-shaped hull pieces $\mathrm{SH}_K(O_i)$
             forming the disjoint proxy.}
    \label{fig:2d_static}
\end{figure}

After building all proxy hulls, any pair of proxies whose pieces intersect are
merged via union-find and the loop repeats. Fig.~\ref{fig:2d_static} illustrates this construction for a cluster of three 2-D ellipse obstacles: the admissible kernel cones (one per obstacle per excluded point) are intersected to yield the feasible kernel region, the selected kernel triangle $K$ is placed at its centroid, and the per-obstacle star-shaped hull pieces $\mathrm{SH}_K(O_i) = \mathrm{CH}(O_i \cup K)$ are assembled into the final disjoint proxy.

\subsection{Reactive Dynamical System Planning}

\subsubsection{Nominal Dynamics}

A goal-attractor DS provides the nominal end-effector velocity.  In practice
the velocity is normalized to a unit vector so that speed is controlled
separately by $v_{\max}$:

\begin{equation}
f_0 = \frac{x_g - x}{\|x_g - x\|}
\end{equation}

\subsubsection{Modulation Matrix}

For each star-world proxy $O'_i$ the controller builds a modulation matrix
$M_i$ that suppresses motion into the proxy while amplifying tangential motion.
The kernel centre $x_{c,i} \in \mathrm{int}\,\ker(O'_i)$ guaranteed by
Algorithm~2 serves as the proxy's reference point~\cite{huber2019avoidance}.
The normalized distance uses the direction-dependent proxy radius $r_i(x)$
(the support function of the proxy hull in the direction $(x - x_{c,i})$):

\begin{equation}
\Gamma_i(x) = \frac{\|x - x_{c,i}\|}{r_i(x) + \delta}
\end{equation}

An orthonormal frame is built from the outward normal and two tangent vectors:

\begin{equation}
E_i(x) = \begin{bmatrix} n_i(x) & e_{t1}(x) & e_{t2}(x) \end{bmatrix},
\qquad n_i(x) = \frac{x - x_{c,i}}{\|x - x_{c,i}\|}
\end{equation}

The eigenvalue matrix encodes the asymmetric response:

\begin{equation}
D_i(x) = \mathrm{diag}\!\left(
    1 - \frac{\eta}{\Gamma_i},\;
    1 + \frac{\eta}{\Gamma_i},\;
    1 + \frac{\eta}{\Gamma_i}
\right)
\end{equation}

Because $E_i$ is orthonormal the modulation matrix is $M_i = E_i D_i E_i^T$.
Multiple proxies are combined via proximity-weighted superposition with weights
$w_i \propto \Gamma_i^{-p}$ (exponent $p = 2$):

\begin{equation}
\dot{x}_{\text{mod}} = \frac{M(x)\,f_0}{\|M(x)\,f_0\|}\,v_{\max},
\qquad M(x) = \sum_i w_i M_i(x)
\end{equation}

\section{Franka Kinematics and Control}

The Franka Panda arm is a 7-DOF serial manipulator with joint vector
$q \in \mathbb{R}^7$.  Forward kinematics map joint angles to end-effector
position, $x = f(q)$, and differentiating gives:

\begin{equation}
\dot{x} = J(q)\,\dot{q}
\end{equation}

where $J(q) \in \mathbb{R}^{3 \times 7}$ is the translational Jacobian of the
end-effector link.  Rather than inverting the Jacobian explicitly, the
implementation uses PyBullet's built-in damped IK solver
(\texttt{calculateInverseKinematics}) seeded with the current joint
configuration $q_{\text{cur}}$ at each control step.  Seeding from the current
state minimizes joint motion and prevents discontinuous solution jumps:

\begin{equation}
q_{\text{IK}} = \mathrm{IK}\!\left(x_{\text{target}},\; q_{\text{cur}}\right)
\end{equation}

where $x_{\text{target}} = x + \dot{x}_{\text{mod}}\,\Delta t$ is the
Cartesian step from the modulated velocity.  Joint limits
$q \in [q_{\min},\, q_{\max}]$ are enforced directly inside the IK call.

\subsection{Null-Space Arm-Body Avoidance}

End-effector modulation alone cannot prevent the arm links from colliding with
obstacles.  A secondary APF layer monitors all nine Franka links.  Each link is
approximated as a capsule sampled at three fractions along the segment between
consecutive joint origins.  When any sample point has $\Gamma < 1.6$, a
repulsion correction $\Delta q_{\text{corr}}$ is computed in joint space via
the Jacobian transpose.

To apply this correction without disturbing the end-effector trajectory, it is
projected into the null space of the EE Jacobian:

\begin{equation}
N = I - J^\dagger J, \qquad J^\dagger = \mathrm{pinv}(J)
\end{equation}

\begin{equation}
\Delta q_{\text{ns}} = N\,\Delta q_{\text{corr}}
\end{equation}

The null-space step is scaled so that $\|\Delta q_{\text{ns}}\| \leq 0.07$\,rad
(scaling the whole vector preserves the null-space direction, whereas
per-component clipping would cause EE drift).  A posture-restoring term
simultaneously biases the arm toward a nominal elbow-up configuration.  Both
corrections are smoothed with an exponential moving average ($\alpha = 0.10$)
to suppress chattering.  The final joint command is:

\begin{equation}
q_{\text{cmd}} = \mathrm{clip}\!\left(q_{\text{IK}} + \Delta q_{\text{ns}},\;
    q_{\min},\, q_{\max}\right)
\end{equation}

Joint commands are sent via PyBullet position control at 40\,Hz while the star
world is rebuilt at 4\,Hz to limit computation overhead.

\section{Implementation Details}

The system is implemented in Python using PyBullet for physics simulation and
visualization of the Franka Panda arm.  The \texttt{starworlds} package provides
Algorithm~2, the reactive controller, and the \texttt{StarWorldUpdater} stateful
wrapper that auto-detects 2-D or 3-D geometry from the obstacle type.

The implementation includes:
\begin{itemize}[leftmargin=*]
    \item Obstacle generation and inflation via \texttt{Ellipsoid} objects,
    \item Star-World reshaping at 4\,Hz (\texttt{create\_star\_world\_3d}),
    \item DS modulation at 40\,Hz (\texttt{ReactiveController}),
    \item Null-space APF arm-body avoidance across all 9 Franka links,
    \item PyBullet IK position control with joint-limit enforcement,
    \item Visualization of proxy hull wireframes and EE trajectory traces,
    \item Evaluation metrics: success rate, path length ratio, and
          star-world update runtime.
\end{itemize}

The full source code is available at \url{https://github.com/saayuj/star-worlds}
and a demonstration video at \url{https://youtu.be/MY4ZFPl7pGw}.

\section{Experimental Evaluation}

\subsection{Simulation Setup}

Experiments are performed in PyBullet using a simulated Franka Panda arm across
the following six named scenarios:
\begin{enumerate}[leftmargin=*]
    \item \textbf{test} -- single sphere directly on the approach path;
          the simplest possible sanity check and easy baseline,
    \item \textbf{overlap} -- two heavily overlapping spheres rendered as
          visual-only bodies in PyBullet, so the arm passes through their
          geometry unless actively deflected by the modulation field;
          the canonical before/after comparison for the merging case,
    \item \textbf{simple} -- 4 obstacles (2 dynamic, 2 small static);
          the two static spheres are placed close enough to merge into one
          star-world proxy, demonstrating online cluster formation,
    \item \textbf{moderate} -- 6 obstacles at medium density; some
          pairs cluster dynamically as obstacles move,
    \item \textbf{corridor} -- narrow passage between wall-like obstacle
          arrays; tests behavior when merging may block the only viable gap,
    \item \textbf{wall} -- three-sphere vertical wall placed across the
          direct path; primary showcase for obstacle merging and off-axis
          kernel placement.
\end{enumerate}

Each scenario is run once with Star-World reshaping enabled
(Algorithm~2) and once in \emph{raw mode}, in which Algorithm~2 is bypassed
and the Huber controller modulates directly against each individual inflated
sphere. This controlled comparison isolates the effect of workspace reshaping
from the remaining implementation choices. Each trial has a 60\,s time limit.
A run is labeled successful when the end-effector reaches within 2\,cm of the
goal by the end of the recorded trial.

Because each scenario--mode pair is evaluated with a single deterministic run,
the results are a preliminary case study rather than a statistical estimate of
the planners' success probabilities. Repeated trials with randomized obstacle
motions and initial configurations are required for inferential comparisons.

\subsection{Evaluation Metrics}

\subsubsection{Goal attainment}

A trial is recorded as successful if $\|x(T) - x_g\| < 0.02$\,m by the
end of the recorded trial:

\begin{equation}
R_{\text{success}} = \frac{N_{\text{success}}}{N_{\text{trials}}}
\end{equation}

\subsubsection{Path length and deviation}

Cumulative EE path length is accumulated at 40\,Hz:

\begin{equation}
L = \sum_{t=1}^{T-1} \|x_{t+1} - x_t\|
\end{equation}

Path deviation measures how much extra distance the planner travels relative
to the straight-line distance covered toward the goal:

\begin{equation}
\Delta L(t) = L(t) - \|x_g - x_0\| \cdot \frac{\text{progress}(t)}{100}
\end{equation}

The path length ratio $L / \|x_g - x_0\|$ provides a scene-normalized
efficiency measure.

\subsubsection{Computational runtime}

Star-world update time and velocity computation time are recorded separately
using wall-clock timestamps at each invocation:

\begin{equation}
T_{\text{update}} = T_{\text{cluster}} + T_{\text{reshape}}, \qquad
T_{\text{vel}} = T_{\text{modulation}}
\end{equation}

Mean values are reported in milliseconds.  The star world runs at 4\,Hz and
the velocity controller at 40\,Hz, so the relevant budgets are 250\,ms and
25\,ms respectively.

\subsection{Results}

\begin{figure*}[t]
    \centering
    \begin{subfigure}{0.48\textwidth}
        \centering
        \includegraphics[width=\linewidth, height=0.9\textwidth]{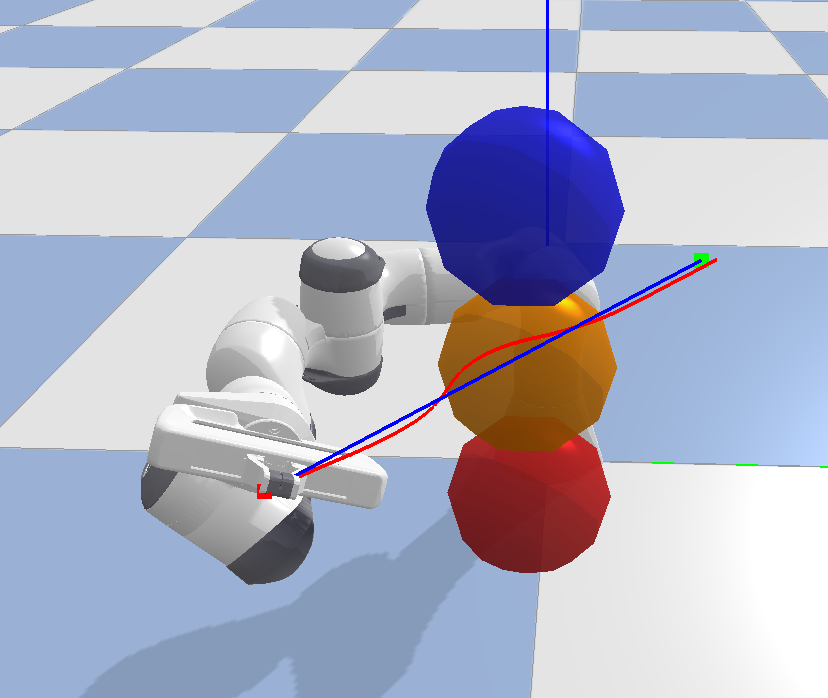}
        \caption{Without Star-World reshaping}
    \end{subfigure}
    \hfill
    \begin{subfigure}{0.48\textwidth}
        \centering
        \includegraphics[width=\linewidth, height=0.9\textwidth]{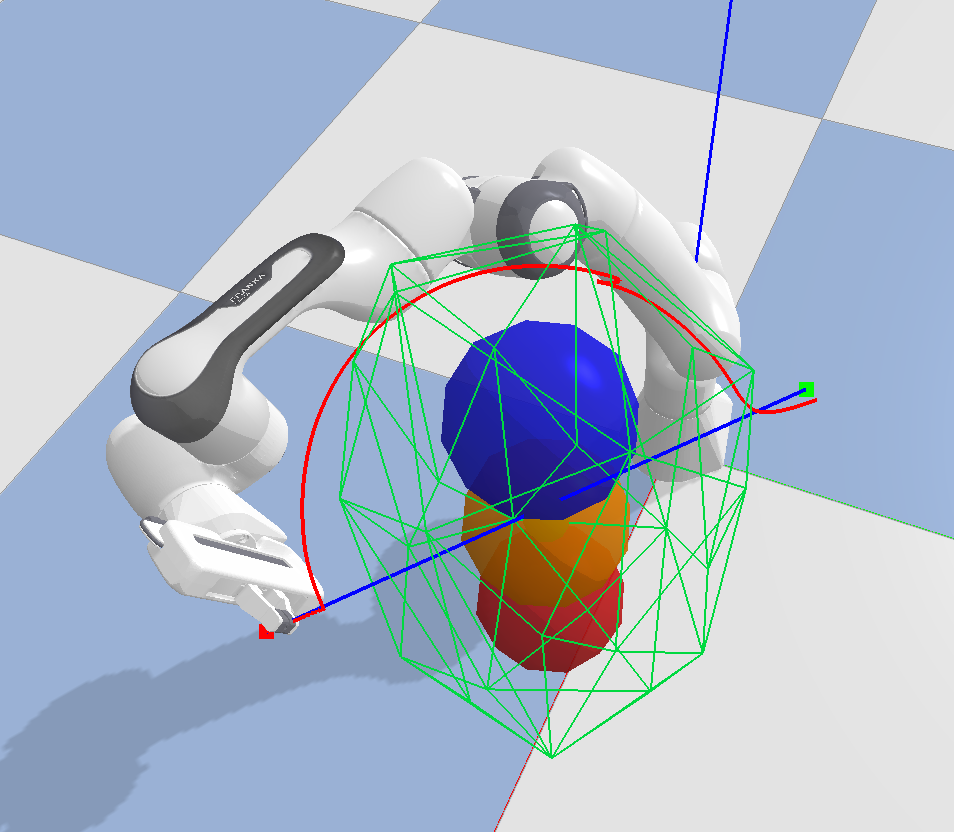}
        \caption{With Star-World reshaping}
    \end{subfigure}
    \caption{\small Wall-scenario comparison between (a) the unreshaped baseline and
             (b) Star-World workspace reshaping.}
\end{figure*}

All six scenarios were run in PyBullet simulation with a 60\,s timeout.  Each trial was run once per mode (raw baseline vs.\ Star-World).
The raw baseline uses the Huber modulation controller directly against one
inflated sphere proxy per obstacle; the Star-World mode first runs
Algorithm~2 to produce a disjoint proxy set and then applies the same
controller to the resulting proxies.

\begin{table*}[t]
\caption{Per-Scenario Results: Raw Baseline vs.\ Star-World Reshaping}
\centering
\begin{tabular}{llcccccc}
\toprule
Scenario & Mode & Success & Path (m) & Straight (m) & Ratio & Sim.\ Time (s)
         & Avg.\ SW Update (ms) \\
\midrule
test      & Raw         & \checkmark & 0.541 & 0.384 & 1.41  & 4.3  & ---  \\
          & Star-Worlds & \checkmark & 1.960 & 0.384 & 5.10  & 60.0 & 0.68 \\
\midrule
overlap   & Raw         & \texttimes & 0.836 & 0.420 & 1.99  & 6.0  & ---  \\
          & Star-Worlds & \checkmark & 2.751 & 0.420 & 6.55  & 60.0 & 1.79 \\
\midrule
simple    & Raw         & \checkmark & 0.846 & 0.843 & 1.004 & 6.5  & ---  \\
          & Star-Worlds & \checkmark & 0.901 & 0.843 & 1.07  & 6.7  & 6.52 \\
\midrule
moderate  & Raw         & \checkmark & 0.919 & 0.910 & 1.01  & 7.0  & ---  \\
          & Star-Worlds & \checkmark & 1.343 & 0.910 & 1.48  & 10.2 & 6.98 \\
\midrule
corridor  & Raw         & \checkmark & 1.019 & 1.040 & 0.98  & 7.8  & ---  \\
          & Star-Worlds & \texttimes & 7.136 & 1.040 & 6.86  & 60.0 & 8.70 \\
\midrule
wall      & Raw         & \texttimes & 0.633 & 0.640 & 0.99  & 5.1  & ---  \\
          & Star-Worlds & \checkmark & 0.954 & 0.640 & 1.49  & 7.3  & 2.68 \\
\bottomrule
\end{tabular}
\end{table*}

\subsubsection{Quantitative summary}

Across the six evaluated scenarios, the raw baseline reaches the goal in
4/6 cases (67\%) with a mean
path-length ratio of $1.10 \pm 0.18$ and a mean completion time of
$6.4 \pm 1.4$\,s among successful runs.  Star-World reshaping reaches the goal in 5/6 cases (83\%); for the three clean convergences (\textit{simple},
\textit{moderate}, \textit{wall}) the mean path-length ratio is
$1.35 \pm 0.21$ and the mean completion time is $8.1 \pm 1.8$\,s.
The two remaining Star-World successes (\textit{test}, \textit{overlap})
converge only at the 60\,s timeout limit with high path ratios (5.10 and
6.55), indicating near-equilibria that are eventually escaped.
Star-World update time scales from 0.68\,ms (1 obstacle) to 8.70\,ms
(7 obstacles), well within the 250\,ms budget at 4\,Hz.  Velocity
computation remains $\approx 0.3$\,ms per step in both modes.

\subsubsection{Scenario analysis}

\textbf{simple and moderate.}  These are the scenarios best suited to
Star-World reshaping.  Obstacles are sufficiently spread that Algorithm~2
produces a small number of well-separated proxies.  In \textit{simple},
two nearby static spheres merge into one proxy; both modes succeed with
essentially identical path lengths (ratio 1.07 vs.\ 1.004).  In
\textit{moderate}, six partly dynamic obstacles are consolidated into a
single large proxy; the Star-World trajectory is longer (ratio 1.48 vs.\
1.01) but converges cleanly in 10.2\,s vs.\ 7.0\,s.  In both cases the
unified proxy provides a consistent repulsion direction and the arm reaches
the goal without oscillation.

\textbf{wall.}  The three-sphere vertical wall is the primary showcase for
obstacle merging.  All three spheres fuse into a single proxy and the kernel
is placed off the approach axis, giving the modulation field a coherent
tangential direction; the Star-World arm arcs cleanly around the wall (ratio
1.49, 7.3\,s).  The raw controller fails (5.1\,s): three independent
modulation matrices produce conflicting repulsions that trap the arm between
the spheres without convergence.

\textbf{test.}  A single sphere directly on the approach path.  The raw
controller deflects around it cleanly (ratio 1.41, 4.3\,s).  The Star-World
arm also reaches the goal but requires the full 60\,s timeout (ratio 5.10).
The kernel is placed on the approach axis, initially generating a modulation
vector that partially opposes the goal direction and driving the arm toward
the workspace boundary; the arm eventually escapes the near-equilibrium and
converges, but the resulting path is highly circuitous.

\textbf{overlap.}  The canonical two-sphere overlapping-wall scenario shows
the intended contrast.  The raw baseline fails (6.0\,s): the competing
outward normals from the two spheres cancel on the approach axis, driving the
arm into the pair without deflecting it around.  Star-World reshaping merges
both spheres into a single proxy with an off-axis kernel, producing a
coherent tangential deflection; the arm converges to the goal at the 60\,s
limit (ratio 6.55), indicating some circuitous routing before final
convergence.

\textbf{corridor.}  The raw controller traverses the narrow gap cleanly
(ratio 0.98, 7.8\,s).  Star-World reshaping fails: Algorithm~2 merges all
seven corridor obstacles (six wall spheres plus the moving blocker) into a
single large proxy that entirely fills the corridor cross-section.  The
resulting proxy has no gap for the EE to pass through, so the modulation
field deflects the arm to the workspace boundary and the trial times out
after circulating for 60\,s with path ratio 6.86.

\subsubsection{Failure-case analysis}

Three qualitatively distinct failure modes emerged while running tests.

\textbf{Near-equilibria near workspace boundaries.}  In the
\textit{test} and \textit{overlap} scenarios the EE is temporarily held
near the workspace boundary before eventually converging, producing highly
circuitous paths (ratios 5.10 and 6.55) and consuming the full 60\,s
budget.  These near-equilibria arise from the interaction between the
modulation field (which pushes the EE away from the proxy boundary) and
the IK solver (which clamps the joint targets to workspace limits); the arm
escapes rather than fully stalling, but convergence is degraded.

\textbf{Over-aggressive merging in corridor-like geometry.}  In the
\textit{corridor} scenario, disjoint obstacles that are individually
navigable become a single impassable proxy after merging.  The current
Algorithm~2 implementation merges clusters whenever their inflated radii
overlap, with no mechanism to preserve passable gaps.  The merged proxy
correctly satisfies the mathematical conditions of a disjoint star world,
but it is topologically more conservative than the original obstacle set.

\textbf{Kernel placement inside small single-obstacle proxies.}  When
Algorithm~2 processes a single isolated obstacle, the admissible kernel
region is the exterior complement of the sphere.  The kernel selection
heuristic (centroid of the feasible polygon) can place the kernel point at
an unfortunate location relative to the goal direction, biasing the
modulation away from the goal rather than around the obstacle.

\subsection{Computational Complexity}

Star-world update cost scales as $O(N^2)$ in the number of obstacles $N$
(one cone intersection per obstacle-excluded-point pair), plus
$O(K \cdot M \cdot n_{\text{approx}})$ for the star-shaped hull and
re-clustering steps ($K$ clusters of mean size $M$,
$n_{\text{approx}} = 16$).  Measured update times of 0.68--8.70\,ms for
1--7 obstacles fit comfortably within the 250\,ms budget at 4\,Hz.
Velocity computation costs $\approx 0.3$\,ms per step in both modes.

\section{Discussion}

\subsection{Interpretation and Validity}
The comparison suggests that reshaping is most useful when overlap creates
conflicting modulation directions, as in the wall scenario. It is not uniformly
beneficial: conservative proxy construction can remove feasible homotopy
classes, and the workspace-level geometric construction does not account for
joint limits or full-arm kinematics.

The evaluation has four important validity limitations. First, each
scenario--mode pair contains only one run, so the reported fractions must not be
interpreted as population success probabilities. Second, all tests are
simulation-only and use spherical obstacle models. Third, the three-dimensional
admissible-kernel construction is an approximation and therefore does not
inherit the complete guarantee of the two-dimensional formulation. Fourth, the
raw and reshaped modes share the same IK and arm-body avoidance layers; failures
can therefore arise from controller coupling rather than workspace reshaping
alone.

The results reveal a nuanced picture.  Star-World reshaping performs as
intended in \textit{simple}, \textit{moderate}, and \textit{wall}: it
consolidates overlapping obstacles into a unified proxy, eliminates
competing modulation contributions, and produces smooth convergent
trajectories.  The algorithm is computationally lightweight and operates
comfortably online.  However, three scenarios expose limitations that have
practical importance.

\textbf{Gap preservation.}  The most actionable improvement would be a
gap-aware merging criterion that declines to merge two clusters when the
free corridor between their inflated boundaries is wider than the robot's
effective cross-section.  This would prevent the corridor failure and
allow the planner to handle narrow-passage environments that are currently
blocked by over-merging.

\textbf{Workspace-aware IK integration.}  The secondary equilibria observed
in \textit{test} and \textit{overlap} are not failures of Algorithm~2 per
se, but of the coupling between the EE modulation field and the 3-D IK
solver near joint limits.  Incorporating workspace-boundary repulsion
directly into the reactive controller, or replacing the Euler-integration
IK target with a Riemannian motion policy that respects joint-limit
geometry, would eliminate these spurious attractors.

\textbf{Kernel selection for single-obstacle clusters.}  For clusters
containing only one obstacle, the feasible kernel region is large and
the current centroid heuristic does not account for the goal direction.
A simple improvement would be to bias the kernel centroid toward the
side of the obstacle that minimizes the angle between the resulting proxy
normal and the goal direction, ensuring that the modulation always deflects
the EE around rather than away from the goal.

\textbf{Dynamic and real-world extensions.}  Incremental star-world
updates (re-clustering only changed obstacles) would reduce latency for
fast-moving scenes.  Deployment on a physical Franka would additionally
require ellipsoid fitting from depth-sensor point clouds and online radius
estimation.

\textbf{Full-arm star world.}  The current approach applies Algorithm~2
only in EE space; arm-body avoidance relies on an APF null-space correction
that does not benefit from workspace reshaping.  Extending the star-world
framework to joint space, or applying it independently at multiple arm
cross-section points, could provide stronger arm-body safety guarantees in
cluttered manipulation tasks.

\textbf{Adaptive star-world activation.}  A practical engineering
consideration emerging from the results is that Star-World reshaping
should not necessarily be applied at all times.  The algorithm introduces
non-trivial computational overhead (up to $\approx$9\,ms per update) and
tends to increase path length in scenarios where obstacles do not
meaningfully overlap.  When obstacles are well-separated and individually
navigable, the raw modulation controller is both faster and more
path-efficient.  A promising direction is therefore to design a switching
mechanism that monitors the current obstacle configuration --- e.g., by
checking whether any pair of inflated obstacle boundaries overlap --- and
activates Algorithm~2 only when overlapping clusters are detected.  This
would preserve the convergence guarantees of star-world reshaping
precisely where they are needed, while falling back to the lighter raw
controller in benign scenes.

\section{Conclusion}

This paper presented a full 3-D reactive planning pipeline for a Franka
Panda arm using Star-World workspace reshaping (Algorithm~2 of Dahlin \&
Karayiannidis~\cite{dahlin2023}).  The pipeline integrates online obstacle
clustering, star-shaped hull proxy construction, Huber-modulation end-effector
control, and null-space APF joint-space arm-body avoidance, running at 4\,Hz
for world updates and 40\,Hz for velocity commands.

Experimental results across six PyBullet scenarios show that Star-World
reshaping achieves a 5/6 success rate vs.\ 4/6 for the raw baseline.
In \textit{simple}, \textit{moderate}, and \textit{wall} it produces clean
convergent trajectories (path-length ratios 1.07--1.49); in \textit{overlap}
it correctly handles the canonical two-sphere wall that defeats the raw
controller.  The star-world update scales from 0.68\,ms to 8.70\,ms for
1--7 obstacles, well within the online planning budget.

Residual limitations include near-equilibria in single-obstacle and
two-sphere scenarios (arm converges only at the 60\,s timeout limit),
over-aggressive merging that blocks navigable corridors, and suboptimal
kernel placement on the approach axis.  These highlight that Algorithm~2's
correctness guarantees apply to the mathematical proxy structure and do not
automatically transfer to a 7-DOF IK-controlled arm.  Future work should
address gap-aware merging, workspace-boundary-aware modulation, and
goal-directed kernel selection.

\section*{Acknowledgment}
We thank Prof. Nadia Figueroa and the University of Pennsylvania MEAM 6230
teaching team for their guidance during the development of this work.

\balance

\end{document}